\documentclass[11pt]{article}

\usepackage[T1]{fontenc}
\usepackage[utf8]{inputenc}
\usepackage{lmodern}

\usepackage[margin=1in]{geometry}
\usepackage{graphicx}
\usepackage{float}
\usepackage{booktabs}
\usepackage{amsmath,amssymb}
\usepackage{url}
\usepackage{hyperref}
\usepackage{authblk}
\usepackage{footmisc}

\hypersetup{
  colorlinks=true,
  linkcolor=blue,
  citecolor=blue,
  urlcolor=blue
}

\title{Event-Driven On-Sensor Locomotion Mode Recognition Using a Shank-Mounted IMU with Embedded Machine Learning for Exoskeleton Control\thanks{%
This work was prepared by the authors in their personal capacities. 
The views expressed are those of the authors and do not necessarily 
represent the views of Align Technology, Inc.}
}

\author[1]{Mohammadsaleh Razmi}
\author[1,*]{Iman Shojaei}

\affil[1]{Align Technology, Inc., San Jose, CA 95134, USA}
\affil[*]{Corresponding author: \texttt{ishojaei@aligntech.com}}

\date{}

\begin{document}
\maketitle

\begin{abstract}
This work presents a wearable human activity recognition (HAR) system that performs real-time inference directly inside a shank-mounted inertial measurement unit (IMU) to support low-latency control of a lower-limb exoskeleton. Unlike conventional approaches that continuously stream raw inertial data to a microcontroller for classification, the proposed system executes activity recognition at the sensor level using the embedded Machine Learning Core (MLC) of the STMicroelectronics LSM6DSV16X IMU, allowing the host microcontroller to remain in a low-power state and read only the recognized activity label from IMU registers.
While the system generalizes to multiple human activities, this paper focuses on three representative locomotion modes—stance, level walking, and stair ascent—using data collected from adult participants. A lightweight decision-tree model was configured and deployed for on-sensor execution using ST MEMS Studio, enabling continuous operation without custom machine learning code on the microcontroller. During operation, the IMU asserts an interrupt  when motion or a new classification is detected; the microcontroller wakes, reads the MLC output registers, and forwards the inferred mode to the exoskeleton controller. This interrupt-driven, on-sensor inference architecture reduces computation and communication overhead while preserving battery energy and improving robustness in distinguishing level walking from stair ascent for torque-assist control.
\end{abstract}

\noindent\textbf{Keywords:} human activity recognition; inertial measurement unit; on-sensor inference; embedded machine learning; decision tree; wearable sensing; lower-limb exoskeleton

\section{Introduction}

Human activity recognition (HAR) has become a key enabling technology for rehabilitation, assistive robotics, and human--machine interaction. For lower-limb exoskeletons, accurate and low-latency recognition of user activity is essential to enable smooth transitions between locomotion modes and to support adaptive, context-aware control strategies~\cite{Dollar2008}. Wearable sensing based on inertial measurement units (IMUs) has been widely adopted for HAR to quantify gait and locomotion in real-world environments, providing practical alternatives to laboratory motion-capture systems~\cite{Tao2012}.

Recent IMU-based HAR systems have demonstrated strong performance across daily and locomotion-related activities, with increasing emphasis on embedded and energy-efficient deployment. Prior work investigated recognition methods and identified key challenges related to robustness, real-world variability, and practical deployment constraints~\cite{Gomaa2023}. TinyML research advanced the deployment of machine learning on ultra-low-power wearable platforms, enabling resource-constrained inference suitable for embedded real-time systems~\cite{Warden2019}.

In assistive wearable robotics, HAR directly supports intent-aware control. Real-time recognition has been demonstrated in exoskeleton platforms that combine IMUs with additional sensors (e.g., encoders) and achieve low-latency inference suitable for online control loops~\cite{Jaramillo2022}. IMU-based activity and payload classification has also been reported for occupational exoskeletons, emphasizing the role of reliable context detection for safety and user comfort~\cite{Pesenti2023}. Despite these advances, many reported systems still rely on continuous streaming of raw IMU data to a host processor or external device, increasing power consumption and system complexity~\cite{Wang2024}.

To reduce host-side computation and communication overhead, sensor-level inference architectures have emerged in which feature extraction and classification are performed directly within the sensing device, enabling label-level communication to the host system. Recent work has demonstrated on-device activity recognition in wearable platforms such as smart earbuds, highlighting the feasibility of embedded inference at the sensor edge~\cite{DeVecchi2024}. Such architectures are particularly attractive for wearable robotic systems, where human activity recognition must operate continuously under strict latency, power, and reliability constraints. In lower-limb exoskeletons, accurate recognition of locomotion context is essential for enabling adaptive assistance and smooth transitions between locomotion modes~\cite{Jaramillo2022,Moreira2022}. Prior studies have shown that real-time activity and locomotion mode recognition directly improves exoskeleton control performance and user interaction in wearable assistive systems~\cite{Pesenti2023,Shin2021}.

Building on this direction, this work presents a sensor-level HAR system based on a single shank-mounted IMU (LSM6DSV16X) that performs real-time activity recognition entirely within the sensing device. The embedded Machine Learning Core executes a lightweight decision-tree classifier, allowing the host microcontroller to access high-level activity labels without continuous raw data processing. In this study, the framework is demonstrated using three representative locomotion modes—stance, level walking, and stair ascent—selected to align with mode-dependent knee torque assistance and to emphasize robust discrimination between locomotion contexts. The proposed on-sensor inference architecture provides a practical pathway toward low-latency, energy-efficient context recognition suitable for integration into lower-limb exoskeletons and next-generation assistive wearable systems.

\section{Materials and Methods}

\subsection{Hardware Setup and Sensor Placement}

\begin{figure}[H]
\centering
\includegraphics[width=0.30\linewidth]{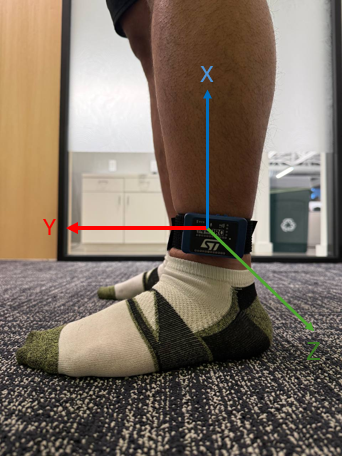}
\caption{Shank-mounted SensorTile.box placement and coordinate definition used for data collection.}
\label{fig:imu_placement}
\end{figure}

The experimental system consisted of an STMicroelectronics SensorTile.box (STEVAL-MKSBOX1V1) equipped with a six-axis LSM6DSV-series inertial measurement unit (IMU) that integrates a 3D accelerometer and a 3D gyroscope. The device was mounted securely on the lateral aspect of the participant's shank using elastic straps to ensure stable placement and to minimize motion artifacts. IMU signals were sampled at highest possible sampling rate of 7.68~kHz, which is sufficient to capture lower-limb dynamics during common locomotion activities, including level walking and stair ascent. Figure~\ref{fig:imu_placement} illustrates the SensorTile.box placement and the coordinate axes used throughout this work.
\subsection{Software and Data Acquisition}

The SensorTile.box platform was configured and operated using ST MEMS Studio, which supports inertial data acquisition, activity labeling, classifier training, and deployment to the embedded Machine Learning Core (MLC) of the IMU.

\begin{figure}[H]
\centering
\includegraphics[width=0.6\linewidth]{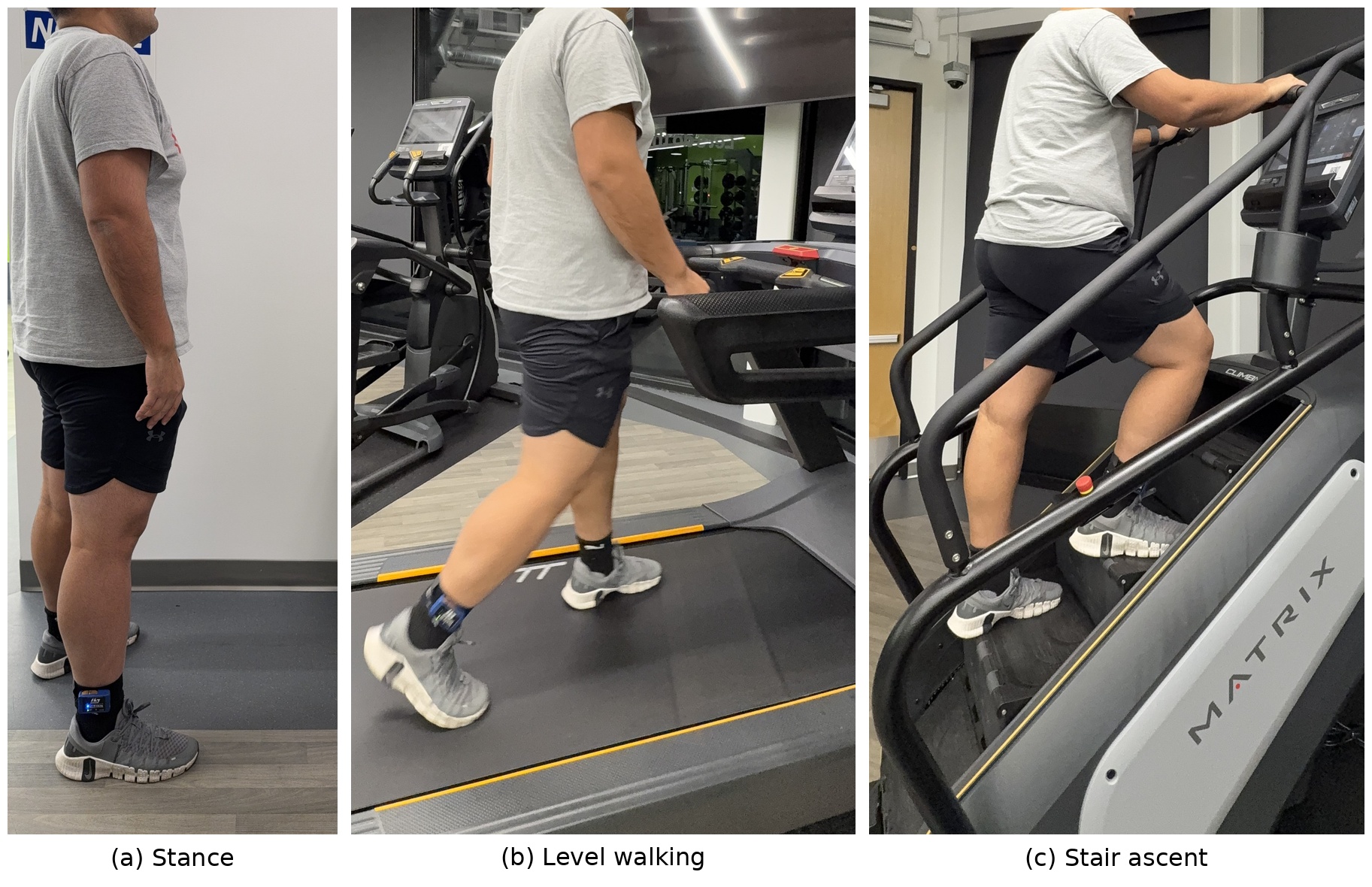}
\caption{Data-collection activities using a shank-mounted IMU: (\textbf{a}) stance, (\textbf{b}) level walking on a treadmill, and (\textbf{c}) stair ascent on a step climber.}
\label{fig:activity_examples}
\end{figure}

Data were collected using a controlled experimental protocol consisting of approximately 2 minutes of level walking on a treadmill and 2 minutes of stair ascent on a step climber, with quiet-standing intervals included to capture stance behavior (Figure~\ref{fig:activity_examples}). During recording, raw accelerometer and gyroscope signals were streamed via Bluetooth Low Energy (BLE) to a host computer for visualization, logging, and offline inspection.

Activity labels corresponding to stance, level walking, and stair ascent were assigned in ST MEMS Studio according to the known protocol segments for each trial. The labeled recordings were subsequently used for feature extraction, post-processing, and decision-tree training.

This work leverages the IMU's embedded Machine Learning Core (MLC) to perform activity recognition directly on the sensor. The trained decision-tree classifier was deployed to the MLC, which generates predicted activity labels exposed through IMU status/output registers. This architecture enables the host microcontroller to log or forward only the final activity label (e.g., to a wearable-robotics controller) without continuous high-rate streaming of raw inertial data.

\subsection{Signal Processing and Activity Classification}

Following labeling, a lightweight decision-tree classifier was trained in ST MEMS Studio and exported as an MLC-compatible configuration. This configuration was then deployed directly to the IMU so that feature extraction and classification executed entirely on-sensor. During real-time operation, the recognized activity label is exposed through dedicated MLC output registers and can be read by the host microcontroller for system-level monitoring and control.
All signal processing, feature extraction, and activity classification were implemented using the embedded Machine Learning Core (MLC) workflow in ST MEMS Studio. Raw inertial data were collected for three classes---stance, level walking, and stair ascent---and exported as CSV files. These labeled CSV recordings were imported into the \textit{Design Pattern} stage of ST MEMS Studio to define the three activity classes and to build the training dataset.

Next, the Activity Feature Selection (AFS) tool was used to configure the accelerometer and gyroscope output data rates (ODR) and to define the MLC inference settings. The MLC ODR was set to 240~Hz with a window length of 240 samples (i.e., a 1~s feature window). Using the expert-mode configuration (including ANOVA-based ranking, AdaBoost and Random Forest evaluations, and Recursive Feature Elimination), the AFS tool identified a compact set of discriminative features suitable for deployment under embedded constraints. In total, 15 features were selected from the inertial channels:
\begin{itemize}
\item $F1$: VARIANCE\_GYR\_X,
$F2$: ENERGY\_GYR\_X,
$F3$: ABS\_MINIMUM\_GYR\_V,
$F4$: MAXIMUM\_GYR\_X,
$F5$: MINIMUM\_ACC\_X,
$F6$: PEAK\_TO\_PEAK\_ACC\_X,
$F7$: MEAN\_GYR\_X,
$F8$: ENERGY\_ACC\_Y,
$F9$: MAXIMUM\_ACC\_X,
$F10$: PEAK\_TO\_PEAK\_GYR\_X,
$F11$: MINIMUM\_GYR\_X,
$F12$: VARIANCE\_ACC\_X,
$F13$: PEAK\_TO\_PEAK\_GYR\_Y,
$F14$: MINIMUM\_GYR\_Y,
$F15$: MEAN\_ACC\_Z.
\end{itemize}

\begin{figure}[t]
  \centering
  \includegraphics[width=\linewidth]{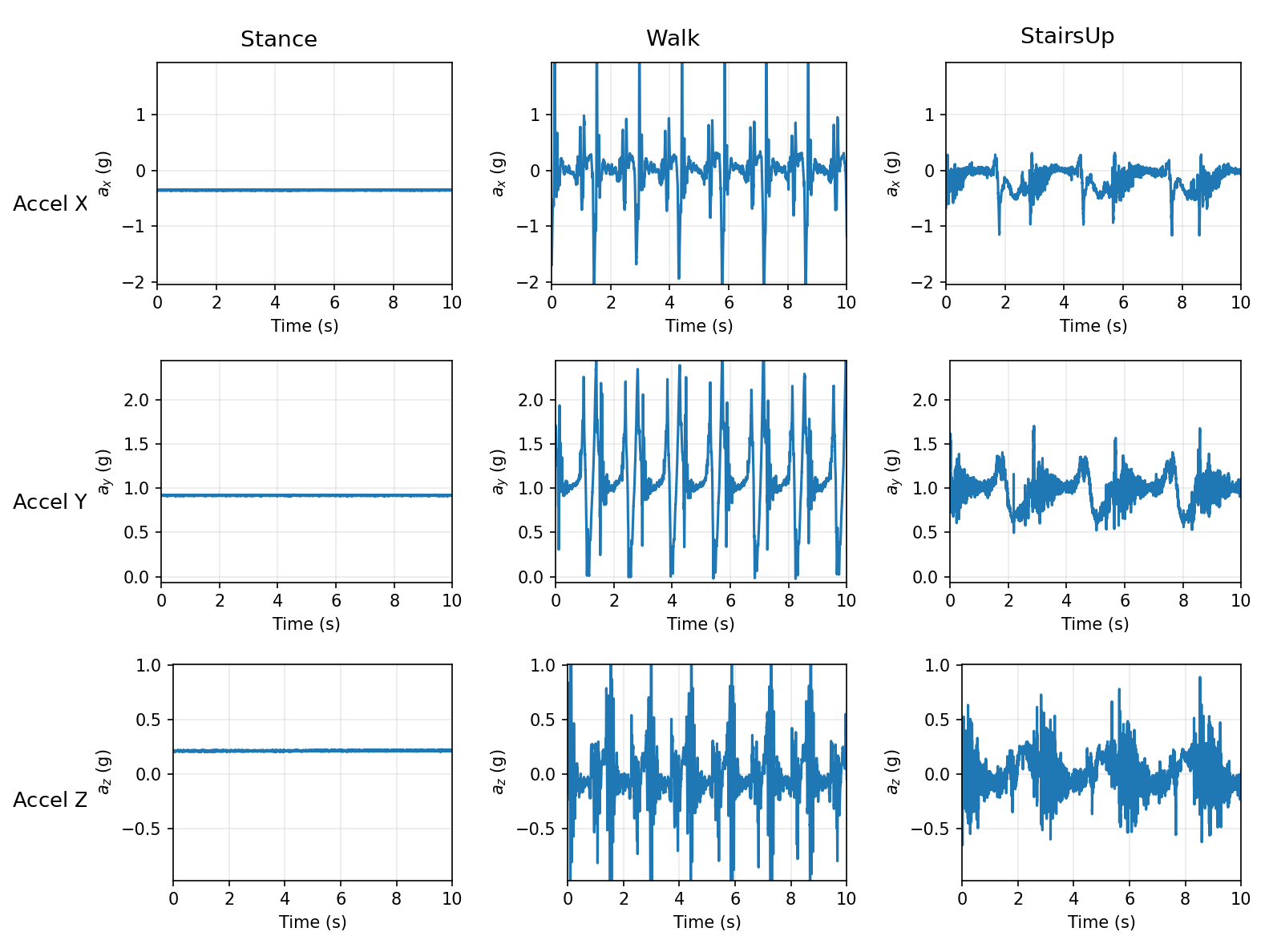}
  \caption{Tri-axial accelerometer signals (10 s) for stance, walking, and stair ascent. Columns: stance, walk, stairsUp. Rows: $a_x$, $a_y$, $a_z$. All subplots in each row share the same y-axis limits for direct comparison.}
  \label{fig:acc_3x3}
\end{figure}

\begin{figure}[t]
  \centering
  \includegraphics[width=\linewidth]{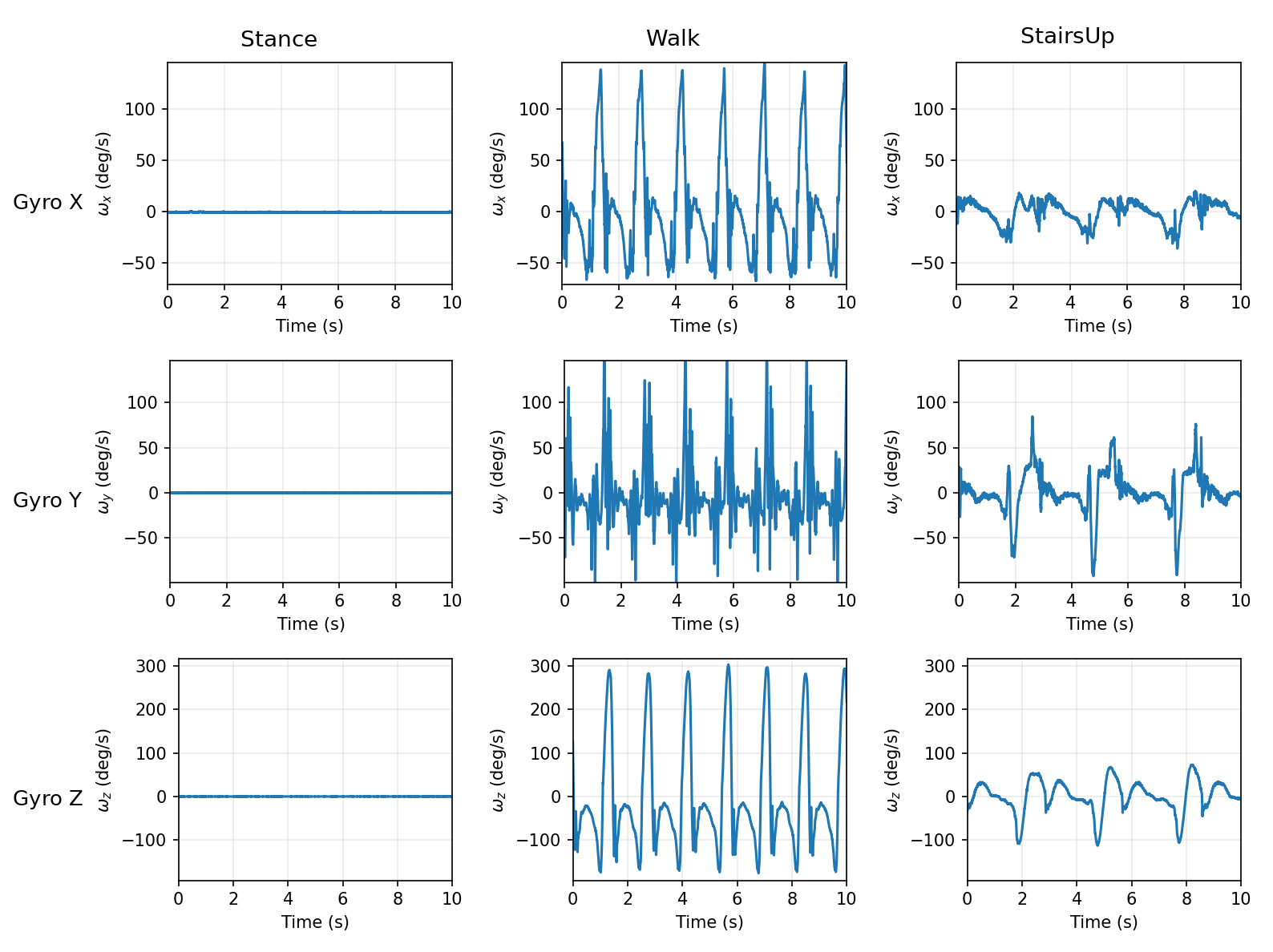}
  \caption{Tri-axial gyroscope signals (10 s) for stance, walking, and stair ascent. Columns: stance, walk, stairsUp. Rows: $\omega_x$, $\omega_y$, $\omega_z$. All subplots in each row share the same y-axis limits for direct comparison.}
  \label{fig:gyr_3x3}
\end{figure}

Representative inertial signal segments for stance, stair ascent, and level walking are shown in Figures~\ref{fig:acc_3x3} and~\ref{fig:gyr_3x3}.

A decision-tree classifier was then generated directly within ST MEMS Studio using the selected features and the labeled dataset. The resulting model was compact (3 leaves, size 5) and produced the three target classes (\textit{walk}, \textit{stairsUp}, \textit{stance}). 
Finally, the trained decision tree was exported as an MLC configuration and deployed to the IMU. During deployment, feature extraction and classification are executed fully on-sensor, and the predicted activity label is reported through the MLC output/status registers. The host microcontroller reads only this label (rather than streaming raw IMU data), enabling low-latency inference with minimal host-side computation.

\paragraph{Training-set performance.}
Using the full labeled dataset for training (43 total instances), the model achieved 100\% accuracy (43/43 correctly classified; Cohen's $\kappa=1.0$). The corresponding confusion matrix was:
\begin{verbatim}
         walk  stairsUp  stance
walk      16      0        0
stairsUp   0     16        0
stance     0      0       11
\end{verbatim}
These results reflect performance on the training set used to generate the decision tree; additional evaluation on held-out sessions/participants is required to quantify generalization under subject and placement variation.
\section{Results}

\subsection{Real-Time Evaluation}
To validate the on-sensor locomotion classifier in a realistic continuous scenario, the trained MLC configuration (selected feature set and decision-tree model) was exported from ST MEMS Studio and programmed into the LSM6DSV16X IMU. The deployed model produced stable locomotion mode estimates with consistent transitions across repeated evaluation trials, demonstrating reliable real-time on-sensor inference. The IMU was worn on the lateral shank as described previously, and the host microcontroller was configured for interrupt-driven operation using INT1 (WAKEUP/MLC events). Upon each interrupt event, the microcontroller woke from a low-power state, read the MLC output/status registers, and logged the predicted locomotion mode. During this evaluation, the host did not stream raw IMU data over BLE and did not perform any signal processing or classification; it acted only as an event-driven reader of the on-sensor inference result. Table~\ref{tab:mode_timeline} summarizes an example timeline extracted directly from the MLC output register, illustrating the temporal consistency of predicted modes during the continuous trial.

\begin{figure}[H]
\centering
\includegraphics[width=0.6\linewidth]{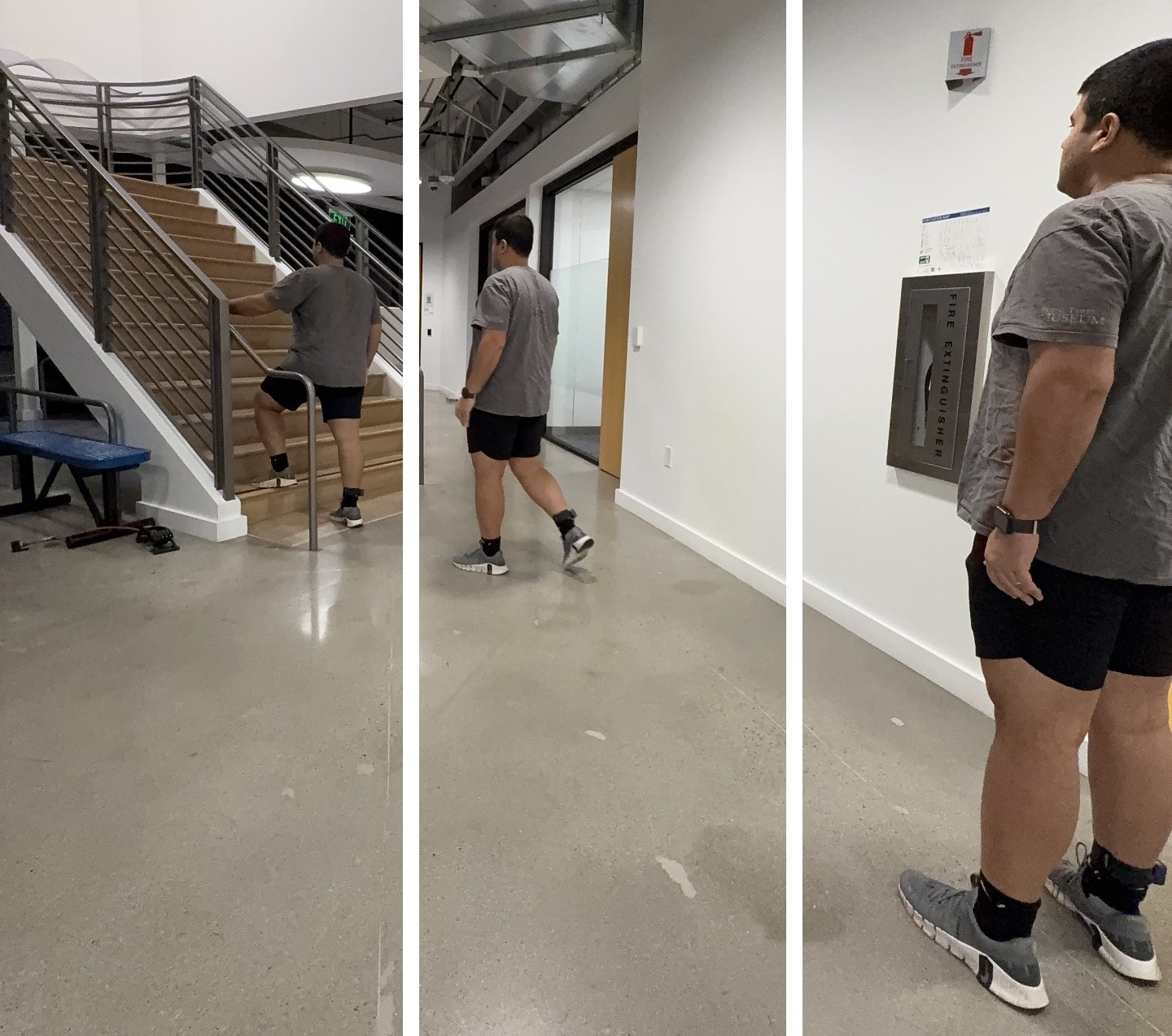}
\caption{Experimental protocol snapshots for real-time evaluation: (\textbf{a}) stance, (\textbf{b}) level walking, and (\textbf{c}) stair ascent. The MLC configuration was pre-loaded on the IMU and the predicted class was read from the MLC output registers during the trial.}
\label{fig:exp_protocol}
\end{figure}
\begin{table}[H]
\caption{Example on-sensor mode timeline from the logged MLC output (\texttt{dec\_tree\_out\_1}).}
\label{tab:mode_timeline}
\centering
\begin{tabular}{lcccc}
\toprule
Mode & Start (s) & End (s) & Duration (s) \\
\midrule
Stance   & 0.00 & 5.39  & 5.39 \\
Walk     & 5.39 & 8.62  & 3.23 \\
StairsUp & 8.62 & 17.24 & 8.62 \\
\bottomrule
\end{tabular}
\end{table}

Each evaluation trial followed a scripted sequence consisting of three consecutive phases: (i) quiet stance, (ii) level walking on a treadmill, and (iii) stair ascent on a step climber. This sequence was designed to emulate common exoskeleton usage transitions (stance $\rightarrow$ walking $\rightarrow$ stair ascent) while remaining short enough to support rapid, repeatable testing.

\paragraph{MLC output encoding.}
The predicted activity label was read from the MLC decision-tree output register (\texttt{dec\_tree\_out\_1}). In the deployed configuration, the three locomotion classes were encoded as stance, walk, and stairsUp. These discrete labels can be forwarded directly to a higher-level exoskeleton controller to select mode-dependent knee torque assistance.

\begin{figure}[H]
\centering
\includegraphics[width=0.90\linewidth]{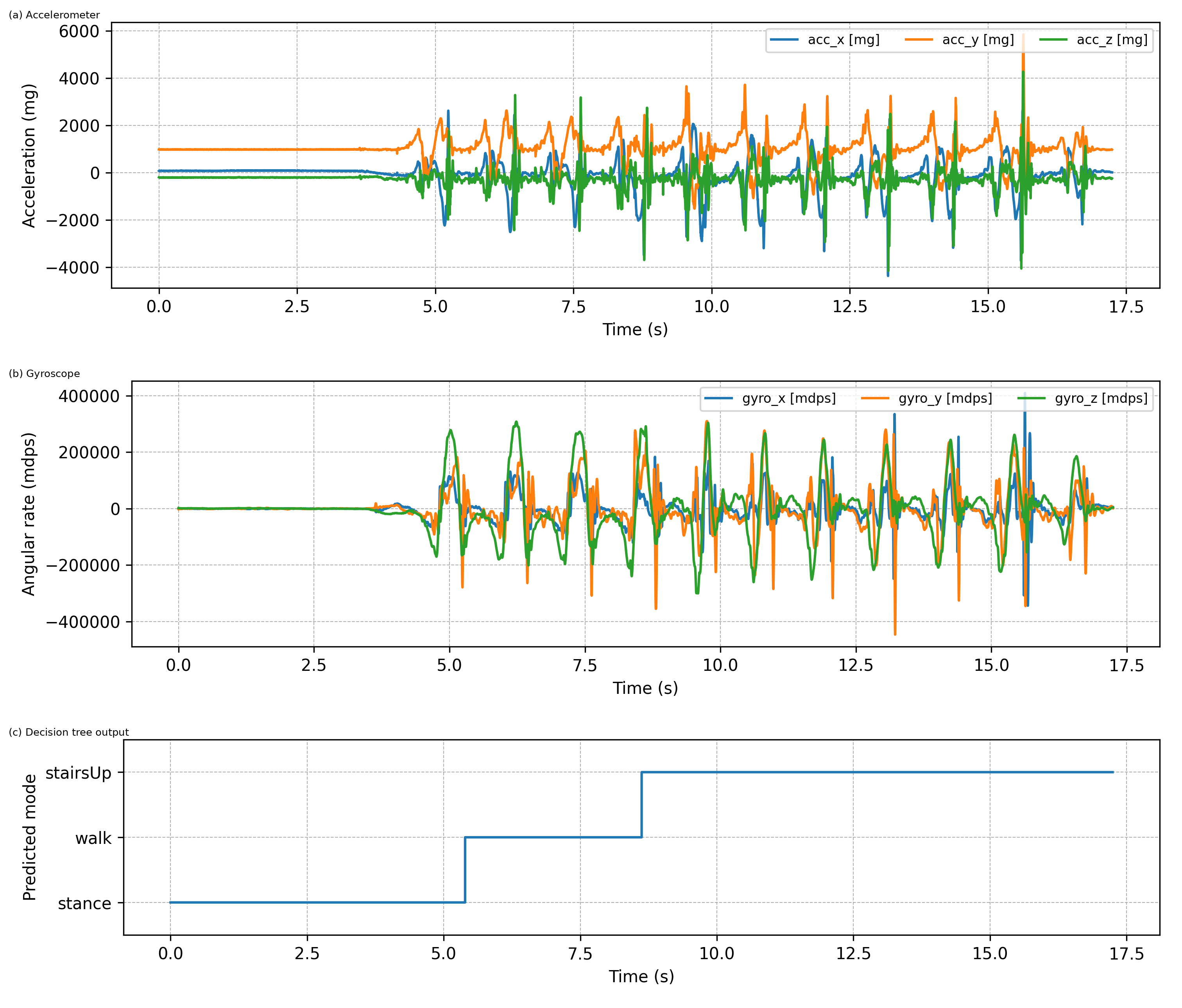}
\caption{Example real-time experiment showing raw IMU signals and on-sensor inference output. (\textbf{a}) Tri-axial accelerometer, (\textbf{b}) tri-axial gyroscope, and (\textbf{c}) MLC decision-tree predicted mode over time (8=stance, 0=walk, 4=stairsUp).}
\label{fig:imu_signals_and_mode}
\end{figure}

\paragraph{Representative real-time inference behavior.}
Figure~\ref{fig:imu_signals_and_mode} shows the trial including tri-axial accelerometer and gyroscope signals along with the corresponding on-sensor decision-tree output. The predicted labels form a stable sequence and indicate two transitions consistent with the intended activity order (stance $\rightarrow$ walk $\rightarrow$ stairsUp). This result demonstrates that the sensor-level classifier can provide low-latency mode information suitable for downstream control while preserving system energy by minimizing host CPU duty cycle and avoiding continuous high-rate IMU streaming. Notably, all classification decisions were generated within the IMU without host-side feature computation or model execution.

\section{Discussion}
This study demonstrates a practical sensor-level activity recognition pipeline for wearable robotics using a single shank-mounted IMU with an embedded MLC. The resulting decision tree is extremely compact and interpretable, which is advantageous for embedded deployment and verification. The learned structure is also physiologically plausible: the peak-to-peak acceleration feature captures the transition from quiet stance to dynamic locomotion, while the gyroscope mean feature reflects differences in shank angular motion patterns between level walking and stair ascent. These observations suggest that simple, well-chosen inertial features can be sufficient to discriminate the locomotion modes most relevant to mode-dependent assistance in lower-limb exoskeletons.

A central contribution of the proposed architecture is the sensor-level inference architecture between the IMU and the host microcontroller. By waking the host only on WAKEUP/MLC events and reading only the final decision label, the system avoids continuous high-rate streaming and reduces host-side computation. This approach is especially attractive for wearable assistive devices where battery lifetime, thermal constraints, and always-on operation are critical. In addition, keeping the classifier within the IMU simplifies integration into embedded control systems by providing a direct mode signal rather than requiring a full host-side sensor-processing pipeline. This result highlights a shift from traditional host-centric HAR pipelines toward sensor-autonomous wearable intelligence.

\paragraph{Limitations and future work.}
The present results should be interpreted in light of several limitations. First, the decision-tree performance reported by ST MEMS Studio reflects the training set used to create the model, and broader validation is needed to quantify generalization across subjects, speeds, and sensor attachment variations. Second, the real-time evaluation reported here demonstrates correct sequencing and stable mode estimates for a representative continuous trial, but additional trials and ground-truth timing (e.g., synchronized video or instrumented devices) would enable formal event-level accuracy, transition delay, and robustness metrics. Future work will focus on (i) multi-subject evaluation with held-out testing, (ii) speed and incline variation, (iii) robustness to sensor placement and strap tightness, (iv) optional inclusion of additional modes if needed (e.g., stair descent) once reliable separation is achieved, and (v) closed-loop integration with an exoskeleton controller to quantify the impact of mode recognition on assistance timing, comfort, and safety.

\section{Conclusions}
This work presented an event-driven, sensor-level locomotion mode recognition system using a single shank-mounted LSM6DSV16X IMU with an embedded Machine Learning Core. A lightweight decision-tree classifier was generated in ST MEMS Studio using automatically computed inertial features and deployed directly to the IMU, enabling fully on-sensor inference without custom machine learning code on the host microcontroller. In a representative real-time trial, the deployed model produced a stable mode sequence (stance $\rightarrow$ walk $\rightarrow$ stairsUp) with clean transitions while the host microcontroller operated as an interrupt-driven label reader. Beyond the demonstrated locomotion modes, the proposed framework establishes a general pathway for embedding machine learning directly within sensing hardware to enable scalable, always-on context awareness in wearable systems. Overall, the proposed approach provides a practical pathway to low-complexity, low-duty-cycle context recognition suitable for wearable assistive robotics and other always-on sensing applications, with future work aimed at broader validation and closed-loop exoskeleton integration.

\section{GenAI Disclosure}
Generative AI tools (ChatGPT, OpenAI) were used to assist with structuring and language refinement of this section. No generative AI was used for data collection, analysis, or result interpretation.





\end{document}